\documentclass[runningheads]{llncs}

\usepackage[T1]{fontenc}
\usepackage{graphicx}
\usepackage{svg}
\usepackage{bbm}
\usepackage{amsmath}
\usepackage{dsfont}
\usepackage{cite}
\usepackage{url}
\usepackage{booktabs}
\usepackage{amssymb}

\begin{document}
\title{WiseLVAM: A Novel Framework For Left Ventricle Automatic Measurements}
\titlerunning{A Novel Framework For Left Ventricle Automatic Measurements}
\author{%
Durgesh Kumar Singh\inst{1}\orcidID{0000-0002-4677-6854} \and
Qing Cao\inst{3} \and
Sarina Thomas\inst{3}\orcidID{0000-0002-1202-0856} \and
Ahc\`ene Boubekki\inst{2}\orcidID{0000-0003-1606-1513} \and
Robert Jenssen\inst{1,4,5}\orcidID{0000-0002-7496-8474} \and
Michael Kampffmeyer\inst{1,4}\orcidID{0000-0002-7699-0405}
}
\authorrunning{Singh et al.}
\author{%
Durgesh Kumar Singh\inst{1}\orcidID{0000-0002-4677-6854} \and
Qing Cao\inst{3} \and
Sarina Thomas\inst{4}\orcidID{0000-0002-1202-0856} \and
Ahc\`ene Boubekki\inst{2}\orcidID{0000-0003-1606-1513} \and
Robert Jenssen\inst{1,5,6}\orcidID{0000-0002-7496-8474} \and
Michael Kampffmeyer\inst{1,5}\orcidID{0000-0002-7699-0405}
}
\authorrunning{Singh et al.}

\institute{%
UiT The Arctic University of Norway, Tromsø, Norway\\
\email{durgesh.k.singh@uit.no, robert.jenssen@uit.no, michael.c.kampffmeyer@uit.no}
\and
Physikalisch-Technische Bundesanstalt (PTB), Braunschweig, Germany\\
\email{ahcene.boubekki@ptb.de}
\and
GE Healthcare, Wuxi, China\\
\email{qing.cao@gehealthcare.com}
\and
GE Healthcare, Oslo, Norway\\
\email{sarina.thomas@gehealthcare.com}
\and
Norwegian Computing Center, Oslo, Norway
\and
Pioneer Centre for AI, Copenhagen, Denmark
}

\maketitle            
\begin{abstract}
Clinical guidelines recommend performing left ventricular (LV) linear measurements in B-mode echocardiographic images at the basal level--typically at the mitral valve leaflet tips--and aligned perpendicular to the LV long axis along a virtual scanline (SL). However, most automated methods estimate landmarks directly from B-mode images for the measurement task, where even small shifts in predicted points along the LV walls can lead to significant measurement errors, reducing their clinical reliability. A recent semi-automatic method, EnLVAM, addresses this limitation by constraining landmark prediction to a clinician-defined SL and training on generated Anatomical Motion Mode (AMM) images to predict LV landmarks along the same. To enable full automation, a contour-aware SL placement approach is proposed in this work, in which the LV contour is estimated using a weakly supervised B-mode landmark detector. SL placement is then performed by inferring the LV long axis and the basal level- mimicking clinical guidelines. Building on this foundation, we introduce \textit{WiseLVAM}-- a novel, fully automated yet manually adaptable framework for automatically placing the SL and then automatically performing the LV linear measurements in the AMM mode. \textit{WiseLVAM} utilizes the structure-awareness from B-mode images and the motion-awareness from AMM mode to enhance robustness and accuracy with the potential to provide a practical solution for the routine clinical application. The source code is publicly available at \url{https://github.com/SFI-Visual-Intelligence/wiselvam.git}.
. 

\end{abstract}

\section{Introduction}
Measuring the left ventricle~(LV) internal dimension~(LVID), interventricular septum~(IVS), and posterior wall~(LVPW) at end-diastole~(ED) and end-systole (ES) is essential for estimating LV volume and mass~\cite{Teichholz1976, Devereux1986}. To standardize these measurements, guidelines~\cite{Lang2015, Mitchell2019} recommend identifying an anchor frame corresponding to ES or ED, and placing a virtual scanline (SL) perpendicular to the LV long axis at the basal level—typically near the mitral valve leaflet tips—in a B-mode echocardiography image acquired from the parasternal long-axis view. Accurate placement of the SL is critical, as measurement reliability depends heavily on anatomical consistency. While the placement at the basal level is considered standard, anatomical variations, such as the existence of the septal bulge, may require slight adjustments of the SL position, for example, moving the SL slightly towards the LV apex to just beyond the septal bulge. This aligns with broader clinical insights~\cite{CardioservLV} indicating that improper SL placement can result in significant errors in LV wall thickness and cavity size estimates,  potentially contributing to diagnostic errors. 
Deep learning methods typically automate LV linear measurements by framing them as a landmark detection task, using heatmap-based localization to estimate key points from B-mode images~\cite{Gilbert2019, Lin2021, Howard2021, Jafari2022, Duffy2022, Wan2023, Mokhtari2023}. However, small errors in peak estimation can cause clinically significant landmark shifts, affecting measurement accuracy. Two main challenges remain: (i) predicted landmarks are not explicitly constrained to the SL, and (ii) models often struggle to perform contour-aware SL placement due to directly regressing LV landmarks from the B-mode input. To address (i), Gilbert et al.\cite{Gilbert2019} combined heatmap and coordinate supervision, while Wan et al.\cite{Wan2023} applied a weighted L1 loss to enhance precision. Mokhtari et al.\cite{Mokhtari2023} found that label smoothing displaces landmarks and proposed a graph-based network to improve accuracy. EnLVAM~\cite{Singha2024EnLVAM} is a semi-automatic framework that enhances landmark prediction reliability by generating and predicting on Anatomical Motion Mode~(AMM) images~\cite{Carerj2003} through the combination of a user-defined SL and B-mode echocardiography video. Landmark predictions are explicitly constrained along this SL, which helps to mitigate over- and underestimation errors caused by misaligned or shifted keypoints. While this design improves accuracy and interpretability, the dependency on manual SL placement limits its scalability and reproducibility.

\begin{figure*}[t]
\centering
\includegraphics[width=\textwidth]{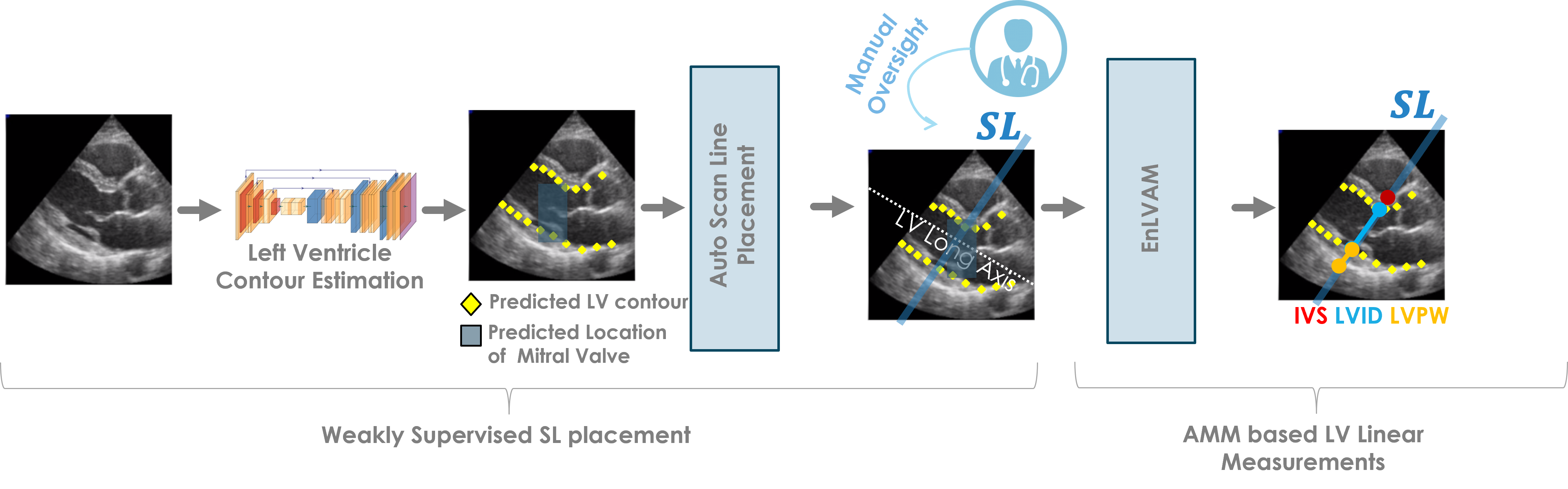}
\caption{
Proposed \textit{WiseLVAM} framework for automatic LV linear measurements.
}
\label{fig:approach}
\end{figure*}

In this work, we extend EnLVAM by introducing \textit{Wise} (Weakly supervised, scanline estimator), a LV contour-aware method for automatic SL placement that eliminates the need for manual input (Figure~\ref{fig:approach}). LV contours are estimated using a weakly supervised B-mode landmark detector, followed by inference of the long axis and basal level to align the SL with true LV geometry. Unlike direct landmark regression, this geometry-driven approach improves accuracy and stability in SL placement, leading to more reliable automatic LV linear measurements when integrated with EnLVAM.
 For training the \textit{Wise}, in order to compensate for the absence of dense contour annotations, a weakly supervised LVID contour generation via EnLVAM is introduced (Figure~\ref {fig:weaksup}). Building on this foundation, we propose \textit{WiseLVAM}—a fully automated yet user-adjustable framework for the automatic LV linear measurement—which integrates the benefits from both B-mode and AMM-based landmark detection and addresses issues (i) and (ii) to improve robustness and accuracy.

\paragraph{The contributions of this work are summarized as follows:}
\begin{itemize}
    
    \item[-] A contour-aware method for automatic  SL placement model~(\textit{Wise}) is introduced, extending EnLVAM by leveraging geometric context to infer the LV long axis and the basal level for anatomically consistent placement.
    
    \item[-] \textit{WiseLVAM}, a fully automated and manually adjustable framework for LV linear measurements, is proposed. It integrates B-mode and AMM-based landmark detection to enhance LV linear measurement accuracy.
    
    \item[-] A novel weakly supervised training strategy is utilized to train the ~\textit{Wise} based on the inferred LVID landmarks from EnLVAM, enabling structure-aware learning without the need for dense manual annotations.
\end{itemize}
\section{Related works}
Landmark detection is typically performed through heatmap regression or classification applied to B-mode frames. In prior work, a U-Net architecture~\cite{Ronneberger2015} was used by Gilbert et al.~\cite{Gilbert2019}, and the DSNT method~\cite{Nibali2018} was integrated to infer landmark coordinates. To improve localization accuracy, a multi-component loss based on heatmap and coordinate errors was proposed. Similarly, atrous convolutions~\cite{Chen2018} were applied by Duffy et al.~\cite{Duffy2022} to enhance LV landmark detection. In another study, Wan et al.~\cite{Wan2023} trained an improved U-Net backbone, introducing a weighted smooth L1 loss to refine the landmark predictions. Despite these advancements, most of these methods rely on heatmap outputs to derive coordinates, often resulting in imperfect localization and subsequent over- or underestimation of LV dimensions. Alternatively, M-mode tracing along a user-defined scanline has been used to visualize cardiac motion in high temporal resolution, facilitating detailed assessment of LV dimensions and wall motion. Although this modality provides a viable route for landmark localization, its automation remains challenging. Accurate M-mode recordings must be acquired with the transducer (M-line) positioned perpendicular to the LV long axis during scanning to prevent oblique measurements. This dependence on operator precision impacts the quality of training data and limits the generalizability of recent M-mode-based approaches~\cite{Tseng2024, Jeong2024}. These limitations motivate the development of a more robust and anatomically consistent approach to automatic LV linear measurement, which we describe in the following section.
\section{Proposed method}\label{sec:method}
This section presents the proposed framework, \textit{WiseLVAM}, as detailed in Sections~\ref{susubsec:wise} and~\ref{susubsec:wiselvam}, beginning with a brief introduction to the landmark detection task in Section~\ref{susubsec:lv_landmark_detection}.

\subsection{LV landmark detection}\label{susubsec:lv_landmark_detection}
Given an input \( m \) containing a single-cycle B-mode echo video \( f_m \), with each frame of size \( H \times W \), and an anchor frame \( \mathbbm{A} \in f_m \) corresponding to the ED/ES phase, the landmark detection model generates softmax-normalized heatmaps \( \hat{H}_i^m \) for each landmark \( i \in N \). The heatmaps have the same resolution as \( \mathbbm{A} \), and are optionally converted to coordinates via the Differentiable Spatial-to-Numerical Transform (DSNT)~\cite{Nibali2018}. The model is trained with a combined loss:
\(
\mathcal{L} = \mathcal{L}_{\text{heatmap}} + \lambda \mathcal{L}_{\text{CE}},
\)
where \( \mathcal{L}_{\text{heatmap}} \) ensures heatmap fidelity and \( \mathcal{L}_{\text{CE}} \) minimizes coordinate error:
\begin{equation*}
\mathcal{L}_{\text{heatmap}} = \frac{1}{N} \sum_{i=1}^{N} \| H_i^{m} - \hat{H}_i^{m} \|^2, \quad
\mathcal{L}_{\text{CE}} = \frac{1}{N} \sum_{i=1}^{N} \| \hat{C}_i^{m} - C_i^{m} \|_2
\end{equation*}

Here, \( H_i^m \) and \( \hat{H}_i^m \) are the ground truth and predicted heatmaps, and \( C_i^m \), \( \hat{C}_i^m \) are the corresponding coordinates.
\subsection{\textit{Wise}: Weakly Supervised Scanline Estimation}\label{susubsec:wise}

Accurate contour-aware automatic placement of the SL is critical for LV linear measurements~\cite{Lang2015, Mitchell2019}. To facilitate this, a weakly supervised approach has been developed for estimating the LV contour and determining anatomically consistent SL position from B-mode images.
\vspace{-0.4cm}
\subsubsection{Training B-mode LV contour estimator}
\begin{figure}[t]
\centering
\includegraphics[width=\textwidth]{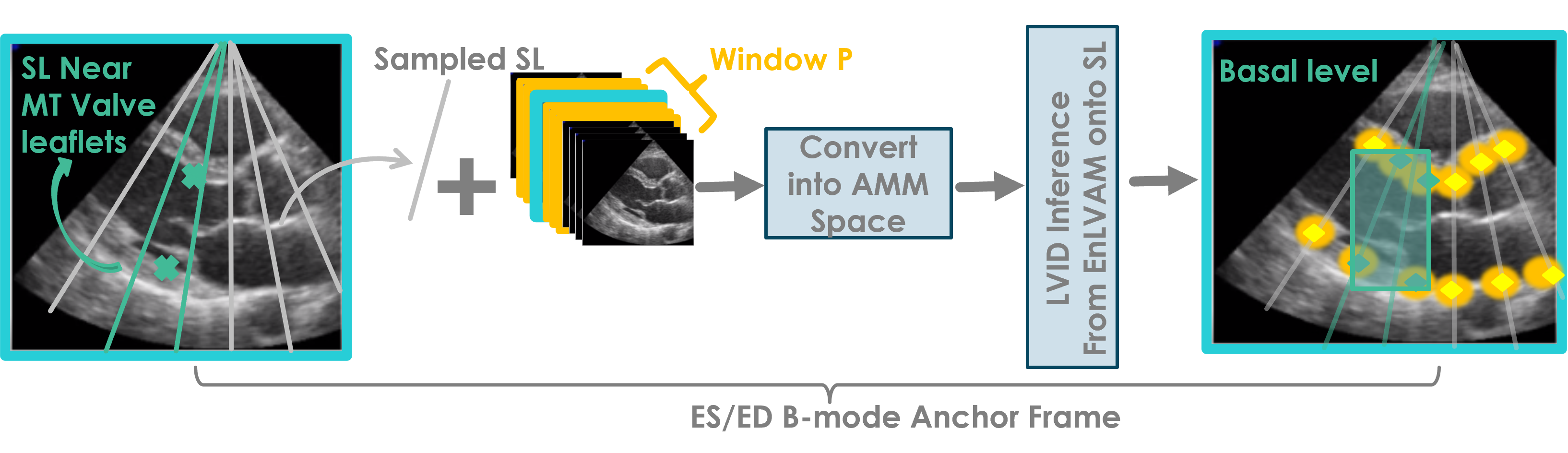}
\caption{Weakly supervised LVID contour generation using EnLVAM~\cite{Singha2024EnLVAM}. $\mathbf{X}$ indicates ground-truth landmarks; $\blacklozenge$ denotes EnLVAM-inferred landmarks with associated uncertainty (ERE).}
\label{fig:weaksup}
\end{figure}

Without ground-truth LV contours, surrogate supervision is employed by sweeping \(N_{\text{LV}}\) SLs across the B-mode anchor frame \( \mathbbm{A} \) (Figure~\ref{fig:weaksup}), and inferring LVID coordinates along each SL using the pre-trained EnLVAM model~\cite{Singha2024EnLVAM}. Two additional SLs, located at the mitral valve leaflet tips to represent the basal level, are obtained from LVID annotations in the training set. In total, this yields \((N_{\text{LV}} + 2) \times 2\) landmarks for training a B-mode-based LV contour estimator. To handle uncertainty in the inferred landmarks, a weak supervision strategy is applied, with uncertainty quantified via the Expected Radial Error (ERE):

\[
\text{ERE}_{i}^{m} = \sum_{(k \in H, l \in W)} \hat{H}_{i,k,l}^m \cdot \| C_{i,k,l}^{m} - \hat{C}_i^m \|_2
\] where \( \hat{H}_{i,k,l}^m \) is the predicted confidence at location \((k, l)\), \( C_{i,k,l}^m \) is the grid coordinate, and \( \hat{C}_i^m \) is the predicted landmark for landmark $i\in N$ and input $m$.  During training, loss contributions for each landmark are weighted inversely by ERE, reducing the influence of uncertain landmarks. This enables learning from weak or noisy annotations without requiring dense contour labels.
\vspace{-0.5cm}
\subsubsection{LV contour-aware scanline placement}
To ensure anatomically consistent measurements, the SL is placed at the basal level and aligned with the LV cavity orientation~(Figure~\ref{fig:approach}). The trained contour estimator predicts \((N_{\text{LV}} + 2) \times 2\) landmarks from the B-mode anchor frame \( \mathbbm{A} \). Midpoints of the \(N_{\text{LV}}\) predicted LVID segments are used to approximate the LV long axis via ridge regression (regularization \( \alpha \)). The SL is then placed at the center of four~(2x2) basal LVID landmarks and oriented perpendicular to the estimated long axis, mimicking clinical guidelines~\cite{Lang2015, Mitchell2019}.

\vspace{-0.1cm}
\subsection{\textit{WiseLVAM}: an automatic yet adaptable framework for LV linear measurements}\label{susubsec:wiselvam}
Building on the contour-aware SL placement strategy (\textit{Wise}), the proposed \textit{WiseLVAM} framework enables fully automatic yet adaptable LV linear measurements. The SL is automatically positioned at the basal level from \textit{Wise}. Clinicians can review and adjust the predicted SLs prior to generating AMM images~(Figure~\ref{fig:approach}), allowing flexibility in cases with anatomical variations (e.g., septal bulge) or low-quality frames where strict automation may be suboptimal.
Once finalized—either automatically or with clinical refinement—the SL is used to extract a virtual AMM image from the B-mode sequence. This image is then processed by the EnLVAM detector, which localizes landmarks along the SL to measure key LV dimensions, including LVID, IVS, and LVPW, with enhanced accuracy and consistency.
\section{Experiments}\label{sec:experiment}
In this section, we cover data processing, experimental setups, and baseline methods adopted in our experiments. 

\paragraph{\textbf{Dataset}}\label{subsec:dataset}
The training dataset extends Gilbert et al.~\cite{Gilbert2019}, consisting of $493$ PLAX echocardiography videos from 306 patients, annotated for ED and ES frames with IVS, LVID, and LVPW measurements, yielding 986 images. It is split as $30\%$ test, $70\%$ train \& validation (approx. 80\% training, 20\% internal validation used for 5-fold cross validation) such that each patient belongs to only one split to avoid data leakage. 

\paragraph{\textbf{Model training \& baselines}} 
Gilbert et al.\cite{Gilbert2019} used a U-Net\cite{Ronneberger2015} combined with a DSNT module~\cite{Nibali2018} for landmark detection, while Wan et al.\cite{Wan2023} introduced an improved variant of U-Net. These models serve as primary baselines and are retrained on the dataset described above, each with their respective loss function. For the proposed scanline estimation model, \textit{Wise}, a vanilla U-Net is trained using an L2 loss. The EnLVAM model\cite{Singha2024EnLVAM} is also trained with U-Net+DSNT, with the multi-component loss in the AMM image space using the same dataset. All models are trained for 60 epochs with a batch size of 2, using 8 gradient accumulation steps and the Adam optimizer~\cite{kingma2014}. A multi-step learning rate schedule is applied, with the initial learning rate $1 \times 10^{-3}$ reduced by a factor of 0.1 at epochs 20 and 40. For \textit{WiseLVAM}, the hyperparameters are set as $\alpha = 1$, $N_{LV} = 20$, and $P = 64$. The AMM model hyperparameter $\lambda$ (used in EnLVAM) is tuned on the validation set.

\paragraph{\textbf{Evaluation metrics}}\label{subsec:evalmetrics}  
Model performance is evaluated using MAE, MAPE, and CE. MAE measures absolute differences in lengths, MAPE normalizes MAE by the ground truth, and CE computes the Euclidean distance between predicted and true coordinates. MAPE is reported in percentage points (pp), MAE and CE in centimeters (cm).
\begin{align*}
\text{MAPE} &= \frac{1}{N-1} \sum_{i=1,j=i+1}^{N-1} \frac{\left| \hat{C}_{ij} - C_{ij} \right|}{C_{ij}}, \quad
\text{CE} = \frac{1}{N} \sum_{i=1}^{N} \left\| \hat{C}_i - C_i \right\|_2, \\
\text{MAE} &= \frac{1}{N-1} \sum_{i=1,j=i+1}^{N-1} \left| \hat{C}_{ij} - C_{ij} \right|
\end{align*}
Where $\hat{C}_{ij}$ and $C_{ij}$ denote the predicted and ground-truth distances between pairs of LV landmarks $i$ and $j$, respectively, and $\hat{C}_i$ and $C_i$ represent the predicted and ground-truth coordinates of the $i$-th LV landmark.

\begin{table}[ht]
\caption{Performance metrics were compared using 5-fold cross-validation. Bold numbers indicate improvements achieved by our approach, bold numbers indicate improvement over both baselines.}
\centering
\begin{tabular*}{\textwidth}{@{\extracolsep{\fill}} llccc}
\toprule
\textbf{Metric} &  \textbf{Structure} & \textbf{WiseLVAM} & \textbf{Gilbert et al.} & \textbf{Wan et al.} \\
 &  & \textbf{(Ours)} & \textbf{\cite{Gilbert2019}} & \textbf{\cite{Wan2023}} \\
\midrule
MAE & IVS    & 0.17$\pm$0.04 & 0.17$\pm$0.03  & 0.18$\pm$0.05 \\
    & LVID   & \textbf{0.24$\pm$0.06}   & 0.32$\pm$0.02  & 0.30$\pm$0.05 \\
    & LVPW   & \textbf{0.15$\pm$0.01}  & 0.19$\pm$0.01  & 0.28$\pm$0.06 \\
    & Overall& \textbf{0.19$\pm$0.03}   & 0.23$\pm$0.01  & 0.25$\pm$0.05 \\

CE  & IVS    & 0.89$\pm$0.19             & 0.82$\pm$0.11   & 0.80$\pm$0.09 \\
    & LVPW   & \textbf{0.93$\pm$0.22}    & 1.22$\pm$0.08  & 1.09$\pm$0.14 \\
    & LVID   & \textbf{0.90$\pm$0.20}    & 1.02$\pm$0.06  & 0.91$\pm$0.08 \\
    & Overall& \textbf{0.90$\pm$0.19}    & 1.02$\pm$0.06  & 0.94$\pm$0.08 \\

MAPE& IVS    & 0.19$\pm$0.04   & 0.19$\pm$0.03  & 0.20$\pm$0.05 \\
    & LVPW   & \textbf{0.16$\pm$0.01}  & 0.20$\pm$0.02  & 0.31$\pm$0.07 \\
    & LVID   & \textbf{0.06$\pm$0.01}   & 0.08$\pm$0.00 & 0.07$\pm$0.01 \\
    & Overall& \textbf{0.13$\pm$0.02}   & 0.16$\pm$0.01  & 0.19$\pm$0.04 \\
\bottomrule
\end{tabular*}
\label{tab:results}
\end{table}
\vspace{-0.2cm}
\section{Results \& discussion}\label{sec:results}
\paragraph{\textbf{Performance evaluation}}
The superior accuracy of \textit{WiseLVAM} in LVPW and LVID measurements, as shown in Table~\ref{tab:results}, can be attributed to contour shape-aware B-mode scanline placement and AMM-constrained landmark predictions from the EnLVAM. This approach ensures more precise measurements, resulting in reduced MAE, MAPE, and CE values.

\paragraph{\textbf{Quality of the predicted SL}}
\begin{table}[t]
\caption{Comparing the SL placement from different methods, with the ground-truth SL in terms of midpoint distance (SL(D), in centimeters) and angle (SL(A), in degrees).}

\centering
\begin{tabular*}{\textwidth}{@{\extracolsep{\fill}} lccc}
\toprule
\textbf{Metric} & \textbf{Wise} & \textbf{Gilbert et al.} & \textbf{Wan et al.} \\
                & \textbf{(Ours)}   & \textbf{\cite{Gilbert2019}} & \textbf{\cite{Wan2023}} \\
\midrule
$SL(D)$ & \textbf{0.76$\pm$0.10} & 0.94$\pm$0.05 & 0.84$\pm$0.12 \\
$SL(A)$ & \textbf{6.07$\pm$1.24} & 8.24$\pm$0.86 & 8.19$\pm$0.94 \\
\bottomrule
\end{tabular*}
\label{tab:ablation}
\end{table}

\begin{figure}[t]
	\centering
	\includegraphics[scale=0.23]
    {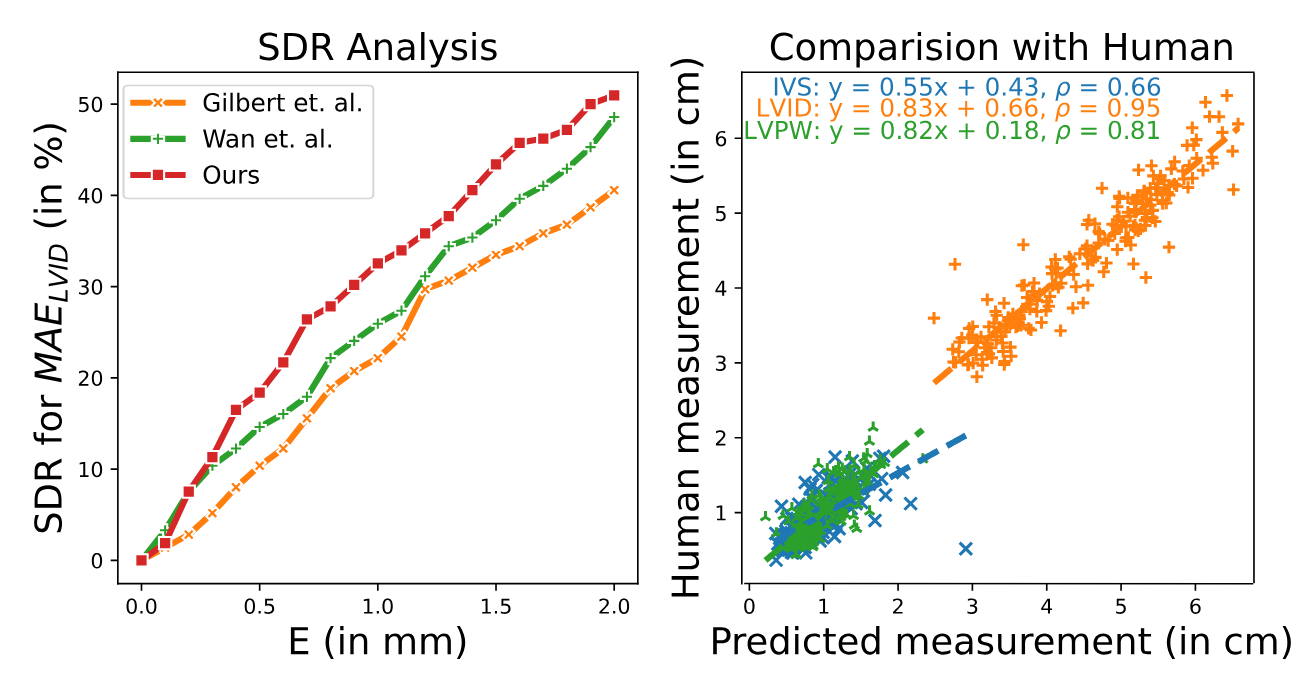}
	\caption{SDR analysis and comparison with human annotator}
	\label{fig:analysis}
\end{figure}

The quality of the contour-aware SL placement is assessed by comparing the predicted SL with the ground-truth SL derived from annotated LVID landmarks in the test data, using angle \(SL(A)\) and midpoint distance \(SL(D)\) as evaluation metrics. For the baseline methods, the SL is inferred from the predicted LVID landmarks. As shown in Table~\ref{tab:ablation}, our method demonstrates higher accuracy across both metrics, indicating improved alignment with expert-defined scanlines resulting in better LV linear measurement estimates.
\paragraph{\textbf{Success Detection Rate for LVID}}  
Success Detection Rate~(SDR) quantifies the proportion of LVID predictions that fall within an acceptable error threshold, making it a clinically relevant metric. Given a test set of size \( |M_{\text{test}}| \) the SDR can be computed as:
\[
\text{SDR} = \frac{1}{|M_{\text{test}}|} \sum_{m=1}^{|M_{\text{test}}|} \mathds{1}[\text{MAE}^{\text{LVID}}_{m} \leq E]
\]
where \( \text{MAE}^{\text{LVID}}_{m} \) is the mean absolute error for LVID measurement in the \( m \)-th test sample, and \( E \) is the predefined threshold (0-2\,mm). As shown in Figure~\ref{fig:analysis}, our method consistently outperforms baselines across error margins up to 2\, mm, highlighting its clinical utility in improving LVID measurement precision.
\paragraph{\textbf{Comparison with human annotator}}
Figure~\ref{fig:analysis} presents the Pearson correlation coefficients~($\rho$) between predicted linear measurements and ground truth annotations for IVS, LVID, and LVPW. Our approach shows the highest correlation for LVID measurements~($\rho = 0.95$), followed by LVPW~($\rho = 0.81$), with the lowest correlation for IVS~($\rho = 0.66$), where the model tends to underestimate. 
\begin{figure}[t]
	\centering
	\includegraphics[scale=0.18]
    {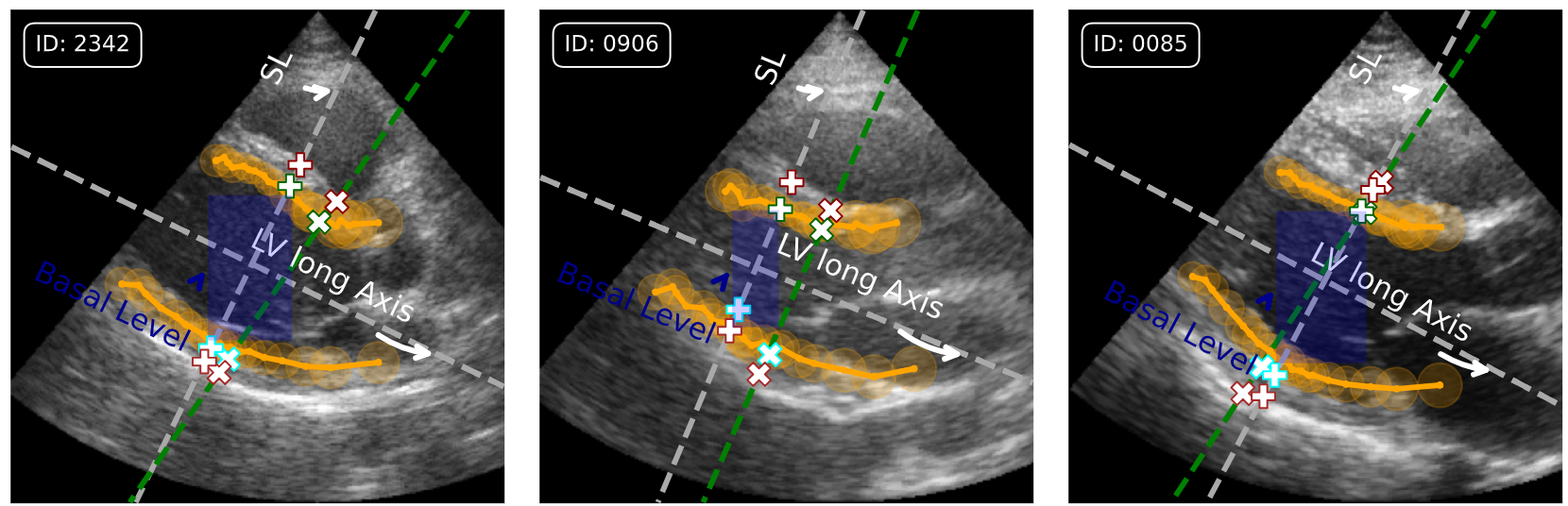}
	\caption{Visualization of Ground Truth~(GT) and predicted landmarks with \textit{WiseLVAM}. (X) denotes GT landmarks, + denotes predicted landmarks along the SL, with the predicted B-mode contour (orange) and basal level (blue). (Best viewed in color)}
\label{fig:visualisation}
\end{figure} 
\paragraph{\textbf{Visualization}}
Figure~\ref{fig:visualisation} illustrates the process of estimating LV linear measurements from B-mode images. The \textit{Wise} model predicts the LV contour using \(N_{\text{LV}} = 20\) landmark segments, identifies the basal level, and estimates the LV long axis. A virtual SL is then placed perpendicular to the estimated long axis, and EnLVAM is used to predict landmarks for automatic LV linear measurement.
\section{Conclusion}\label{sec:conclusion}

In this work, \textit{WiseLVAM}, a novel framework for LV linear measurements, is proposed. A contour-aware, weakly supervised scanline estimation model—\textit{Wise} is introduced to enable anatomically consistent SL placement by learning true LV geometry. The framework integrates B-mode and AMM-based landmark detection in a fully automated yet manually adjustable manner, improving measurement accuracy and demonstrating strong potential for routine clinical application.

\section*{Acknowledgments}
This work was supported by The Research Council of Norway through Visual Intelligence (grant no. 309439), FRIPRO (grant no. 315029), and IKTPLUSS (grant no. 303514).


\end{document}